\documentclass[conference]{IEEEtran}
\ifCLASSINFOpdf
\else
\fi
\usepackage{xspace}
\usepackage{graphicx}
\usepackage{multirow}
\usepackage{booktabs}
\usepackage{hyperref} 
\hypersetup{
    colorlinks=true,
    linkcolor=blue,
    filecolor=magenta,      
    urlcolor=cyan,
}

\newcommand{\LIB}{OSL-ActionSpotting\xspace}
\newcommand{\mysection}[1]{\vspace{2pt}\noindent\textbf{#1}}
\newcommand{\etal}{\textit{et. al.}\xspace}
\newcommand{\ie}{\textit{i.e.}\xspace}
\newcommand{\eg}{\textit{e.g.}\xspace}

\newcommand{\comma}{\,,}

\usepackage{xspace}

\begin{document}
%
% paper title
% Titles are generally capitalized except for words such as a, an, and, as,
% at, but, by, for, in, nor, of, on, or, the, to and up, which are usually
% not capitalized unless they are the first or last word of the title.
% Linebreaks \\ can be used within to get better formatting as desired.
% Do not put math or special symbols in the title.
% \title{OpenSportsLib-AS: A Unified Library\\for Action Spotting in Sports Videos}
\title{\LIB: A Unified Library\\for Action Spotting in Sports Videos}
% \title{OpenSportsLib-AS: A Unified Framework\\for Action Spotting in Sports}
% \title{OpenSportsLib-AS: A Comprehensive Library\\for Action Spotting in Sports}

% author names and affiliations
% use a multiple column layout for up to three different
% affiliations
\author{
\IEEEauthorblockN{Yassine Benzakour}
\IEEEauthorblockA{Montefiore Institute, University of Li\`ege\\Email: yassine.benzakour@student.uliege.be}
\and
\IEEEauthorblockN{Bruno Cabado}
\IEEEauthorblockA{Cinfo \& CITIC Research Center,\\ University of A Coruña\\Email: bruno.cabado@udc.es}
\and
\IEEEauthorblockN{Silvio Giancola}
\IEEEauthorblockA{IVUL, KAUST\\Email: silvio.giancola@kaust.edu.sa}
\and
\IEEEauthorblockN{Anthony Cioppa}
\IEEEauthorblockA{Montefiore Institute, University of Li\`ege\\Email:anthony.cioppa@uliege.be}
\and
\IEEEauthorblockN{Bernard Ghanem}
\IEEEauthorblockA{IVUL, KAUST\\Email: bernard.ghanem@kaust.edu.sa}
\and
\IEEEauthorblockN{Marc Van Droogenbroeck}
\IEEEauthorblockA{Montefiore Institute, University of Li\`ege\\Email: m.vandroogenbroeck@uliege.be}
}

% conference papers do not typically use \thanks and this command
% is locked out in conference mode. If really needed, such as for
% the acknowledgment of grants, issue a \IEEEoverridecommandlockouts
% after \documentclass

% for over three affiliations, or if they all won't fit within the width
% of the page, use this alternative format:
% 
%\author{\IEEEauthorblockN{Michael Shell\IEEEauthorrefmark{1},
%Homer Simpson\IEEEauthorrefmark{2},
%James Kirk\IEEEauthorrefmark{3}, 
%Montgomery Scott\IEEEauthorrefmark{3} and
%Eldon Tyrell\IEEEauthorrefmark{4}}
%\IEEEauthorblockA{\IEEEauthorrefmark{1}School of Electrical and Computer Engineering\\
%Georgia Institute of Technology,
%Atlanta, Georgia 30332--0250\\ Email: see http://www.michaelshell.org/contact.html}
%\IEEEauthorblockA{\IEEEauthorrefmark{2}Twentieth Century Fox, Springfield, USA\\
%Email: homer@thesimpsons.com}
%\IEEEauthorblockA{\IEEEauthorrefmark{3}Starfleet Academy, San Francisco, California 96678-2391\\
%Telephone: (800) 555--1212, Fax: (888) 555--1212}
%\IEEEauthorblockA{\IEEEauthorrefmark{4}Tyrell Inc., 123 Replicant Street, Los Angeles, California 90210--4321}}

% use for special paper notices
%\IEEEspecialpapernotice{(Invited Paper)}

% make the title area
\maketitle

% As a general rule, do not put math, special symbols or citations
% in the abstract
\begin{abstract}
% Background/Introduction:
Action spotting is crucial in sports analytics as it enables the precise identification and categorization of pivotal moments in sports matches, providing insights that are essential for performance analysis and tactical decision-making.
The fragmentation of existing methodologies, however, impedes the progression of sports analytics, necessitating a unified codebase to support the development and deployment of action spotting for video analysis.
% Aim/Objective:
In this work, we introduce \emph{\LIB}, a Python library that unifies different action spotting algorithms to streamline research and applications in sports video analytics.
% Methods:
\emph{\LIB} encapsulates various state-of-the-art techniques into a singular, user-friendly framework, offering standardized processes for action spotting and analysis across multiple datasets. 
% Results:
We successfully integrated three cornerstone action spotting methods into \LIB, achieving performance metrics that match those of the original, disparate codebases. This unification within a single library preserves the effectiveness of each method and enhances usability and accessibility for researchers and practitioners in sports analytics.
% Impact:
By bridging the gaps between various action spotting techniques, \LIB significantly contributes to the field of sports video analysis, fostering enhanced analytical capabilities and collaborative research opportunities. The scalable and modularized design of the library ensures its long-term relevance and adaptability to future technological advancements in the domain.
\end{abstract}

\mysection{Keywords.} Video understanding, Action spotting, Sports Analytics, Python library, Benchmark, Algorithms.

% no keywords

% For peer review papers, you can put extra information on the cover
% page as needed:
% \ifCLASSOPTIONpeerreview
% \begin{center} \bfseries EDICS Category: 3-BBND \end{center}
% \fi
%
% For peerreview papers, this IEEEtran command inserts a page break and
% creates the second title. It will be ignored for other modes.
\IEEEpeerreviewmaketitle

\section{Introduction}

Action spotting in sports videos is a critical task in sports analytics, allowing for the detailed analysis of player actions, game events, and overall performance. 
This process is instrumental in enhancing strategies, training programs, and audience engagement. 
However, the diversity of action spotting methodologies has led to a fragmented landscape where integrating these techniques is challenging.
Existing systems often operate in isolation, leading to inefficiencies and difficulties in achieving a comprehensive analytical perspective. 
The necessity for consistency, efficiency, and scalability in sports analytics drives the need for a unified approach in action spotting. 
Moreover, the rapid evolution of video analysis technologies demands a flexible and adaptable framework that can incorporate new methods and adapt to changing analytical requirements.

\begin{figure}
    \centering
    \includegraphics[width=\linewidth]{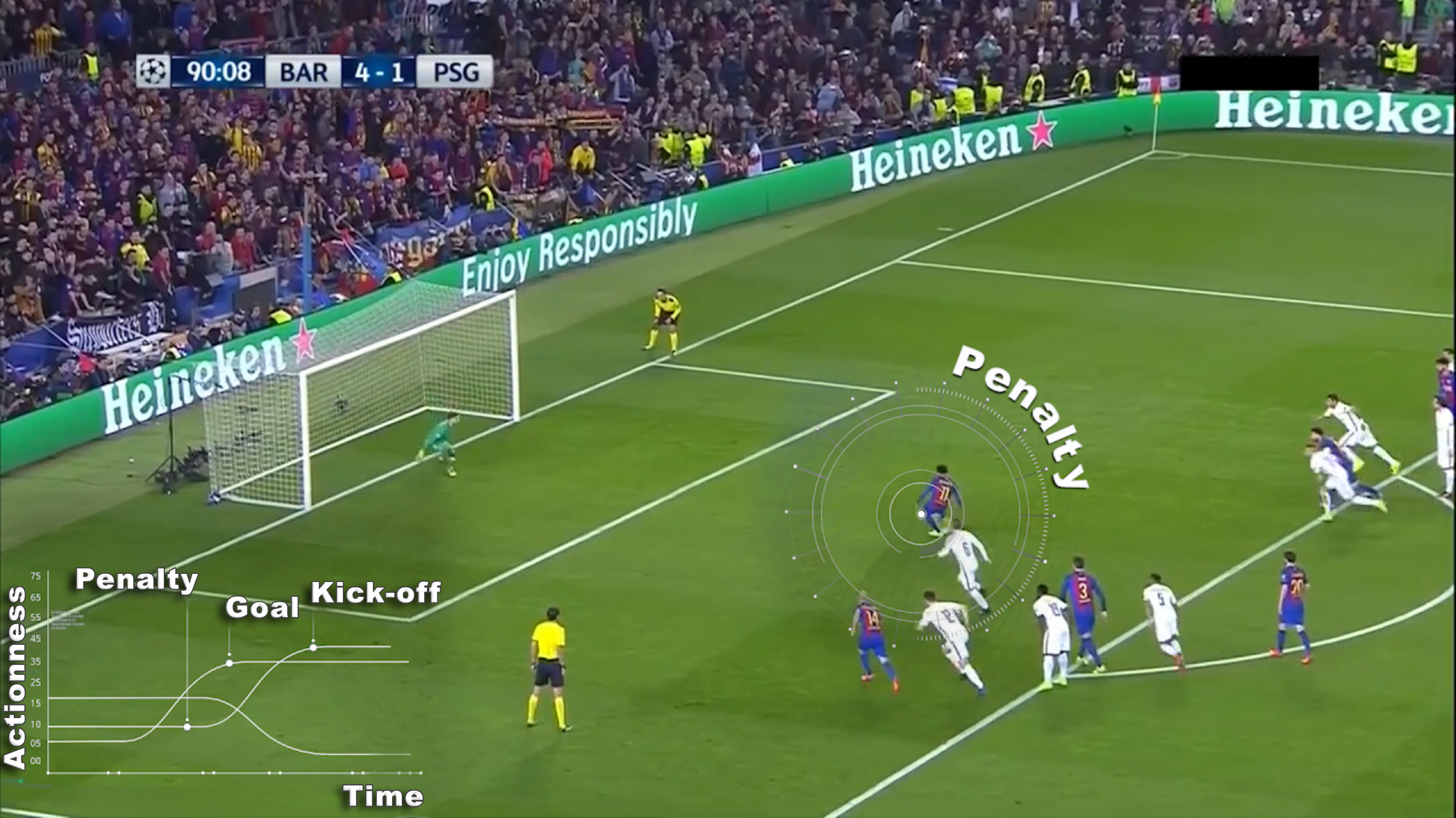}
    \caption{
    \textbf{\LIB} is a plug-and-play library that unifies action spotting algorithms. 
    The design of \LIB is inherently versatile, making it applicable to a broad spectrum of sports video analyses. This adaptability ensures that \LIB can facilitate the development of novel action spotting techniques, 
    and accelerate the deployment of these methods, providing a robust and comprehensive tool for researchers and analysts in various sports domains.
    % \textbf{\LIB} is a plug-and-play library that unifies the algorithms for action spotting.
    % The library is agnostic of the application field, empowering action spotting for any sports video analysis.
    }
    \label{fig:GraphicalAbstract}
\end{figure}

In response to these challenges, we propose \LIB, a comprehensive library that integrates multiple state-of-the-art action spotting methods into a single, cohesive framework, illustrated in Figure~\ref{fig:GraphicalAbstract}. 
This library aims to standardize action spotting across various sports disciplines and enhances the accessibility and efficiency of these analyses for researchers and practitioners alike.
The library is available at 
{\hyperlink{https://github.com/OpenSportsLab/OSL-ActionSpotting}{\underline{https://github.com/OpenSportsLab/OSL-ActionSpotting}}}.
% \hyperlink{https://github.com/OpenSportsLab/OSL-ActionSpotting}{https://github.com/OpenSportsLab/OSL-ActionSpotting}.
% We will publicly release \LIB after the review process.

\mysection{Contributions:}
\textbf{(i)} We release the first action spotting Python library that includes the most impactful algorithms developed in the literature, into a modularized framework that will empower future algorithmic development.
\textbf{(ii)} We develop a video data loader based on DALI, that efficiently pre-process videos to be fed into our algorithm, enabling faster training of end-to-end algorithms.
\textbf{(iii)} We formalize a novel action spotting dataset format in a unique JSON that enables the usage of \LIB's algorithms on new video datasets.
% \textbf{(i)} \LIB represents a significant advancement in the field of sports analytics. By consolidating diverse action spotting algorithms, it provides a standardized and scalable platform that facilitates robust and accurate analysis of sports videos. This unification allows for comparative studies and methodological enhancements, contributing to the ongoing development of sports analytics technologies.
% \textbf{(ii)} Furthermore, the library's design prioritizes user accessibility and computational efficiency, ensuring that it remains a valuable tool for both academic research and practical applications in sports analysis. The integration of five cornerstone action spotting methods and their maintained performance efficacy exemplifies the library's capability to serve as a foundational tool in sports video analytics.

\section{Related Work}

\subsection{Sports Video Understanding}

The field of sports video analysis has evolved into an important research area due to the intricate nature of sports video understanding~\cite{Thomas2017Computer, Naik2022AComprehensive}. Early approaches primarily focused on video classification tasks~\cite{Wu2022ASurvey-arxiv}, which included recognizing distinct actions~\cite{Cioppa2018ABottomUp} and categorizing various phases of gameplay~\cite{Cabado2022Realtime}. Research expanded into different domains of sports video understanding, covering areas such as player detection~\cite{Vandeghen2022SemiSupervised}, tracking~\cite{Maglo2022Efficient}, image segmentation~\cite{Cioppa2019ARTHuS}, and player identification~\cite{Somers2023Body}. It also explored the analysis of tactics ~\cite{Suzuki2019Team}.
%in sports, including football and fencing~\cite{Suzuki2019Team, Zhu2022FenceNet}. 
It investigated specific aspects like pass completion likelihood~\cite{ArbuesSanguesa2020Using}, 3D ball positioning in basketball~\cite{VanZandycke20223DBall}, and the reconstruction of shuttle trajectories in badminton videos~\cite{Liu2022MonoTrack}.
To facilitate research in this domain, numerous research collectives have made available comprehensive datasets, including those by Pappalardo \etal~\cite{Pappalardo2019Apublic}, Yu \etal~\cite{Yu2018Comprehensive}, SoccerTrack~\cite{Scott2022SoccerTrack}, SoccerDB~\cite{Jiang2020SoccerDB}, and DeepSportRadar~\cite{Istasse2023DeepSportradarv2}. The SoccerNet dataset, introduced by Giancola \etal~\cite{Giancola2018SoccerNet}, has set benchmarks for over $12$ specific tasks in soccer video analytics. These include areas such as action spotting~\cite{Deliege2021SoccerNetv2}, camera calibration~\cite{Cioppa2022Scaling, Magera2024AUniversal}, player tracking~\cite{Cioppa2022SoccerNetTracking} and re-identification~\cite{Cioppa2022Scaling}, multi-view foul detection~\cite{Held2023VARS}, comprehensive video captioning~\cite{Mkhallati2023SoccerNetCaption}, explainability~\cite{Held2024XVARS}, depth estimation~\cite{Leduc2024SoccerNetDepth} and game state reconstruction~\cite{Somers2024SoccerNetGameState}. These datasets have become a cornerstone for yearly competitions~\cite{Giancola2022SoccerNet,Cioppa2023SoccerNetChallenge-arxiv}, promoting cooperative research efforts in the field of sports video analysis. 
In this work, we propose an easy-to-use library for action spotting, \ie the task of localizing events in untrimmed videos, identified with a single timestamp.

\subsection{Action spotting} 

Giancola~\etal~\cite{Giancola2018SoccerNet} introduced the task of action spotting, defined as the process of localizing key actions within long untrimmed videos, such as penalties, goals, or corner kicks in football. 
This task differs from temporal activity localization, where activities are identified over durations~\cite{Caba2015ActivityNet}.
In contrast, action spotting focuses on the precise moment of an event, with a single timestamp, in accordance with football regulations~\cite{IFAB2022Laws}. 
Giancola \etal~\cite{Giancola2018SoccerNet} propose a first baseline to solve this task based on learnable pooling methods, later improved with temporally-aware pooling~\cite{Giancola2021Temporally}.
Rongved \etal~\cite{Rongved2021Automated} developed a method using 3D ResNet on video frames analyzed in a sliding window of five seconds. 
Multimodal techniques were explored by Vanderplaetse \etal~\cite{Vanderplaetse2020Improved} and Xarles \etal~\cite{Giancola2022SoccerNet}, who merged visual and auditory data for action detection. 
Further advancements include Cioppa \etal's~\cite{Cioppa2020AContextaware,Cioppa2021Camera} context-aware loss function that leverages temporal dynamics, Vats \etal's~\cite{Vats2020Event-arxiv} multi-tower CNN for managing action localization uncertainty, and Tomei \etal's~\cite{Tomei2021RMSNet} approach of refining feature extraction with a masking strategy targeting post-action frames.
The current state-of-the-art on SoccerNet-v2 for published methods has been set by Denize~\etal~\cite{Denize2024COMEDIAN}, with their end-to-end methodology.
This method outperforms previous leaders like Soares \etal~\cite{Soares2022Temporally}, who employed an anchor-based strategy, and Hong \etal~\cite{Hong2022Spotting}, pioneers of the precise temporal spotting (PTS) technique. 
The PTS approach integrates an end-to-end trainable feature extraction and spotting mechanism, utilizing a lightweight RegNet architecture enhanced with GSM~\cite{Sudhakaran2020GateShift} and GRU~\cite{Cho2014OnThe} modules. 
This surpassed the 2022 challenge's top contenders, showcasing advancements over earlier methods like spatio-temporal encoders~\cite{Darwish2022STE}, graph-based processing layers~\cite{Cartas2022AGraphbased}, and transformer models~\cite{Zhu2022ATransformerbased}.
In this work, we re-implement feature-based and end-to-end methods from cornerstone action spotting algorithms~\cite{Cioppa2020AContextaware,Giancola2021Temporally,Hong2022Spotting,Cabado2024Beyond} into a unified library.
\section{\LIB}

\LIB is a streamlined, plug-and-play library tailored for efficient action spotting in sports videos, suitable for football and other sports. 
It merges various advanced action spotting algorithms into one framework, promoting modularity and facilitating further research development in this field.

\begin{figure*}
    \centering
    \includegraphics[width=\linewidth]{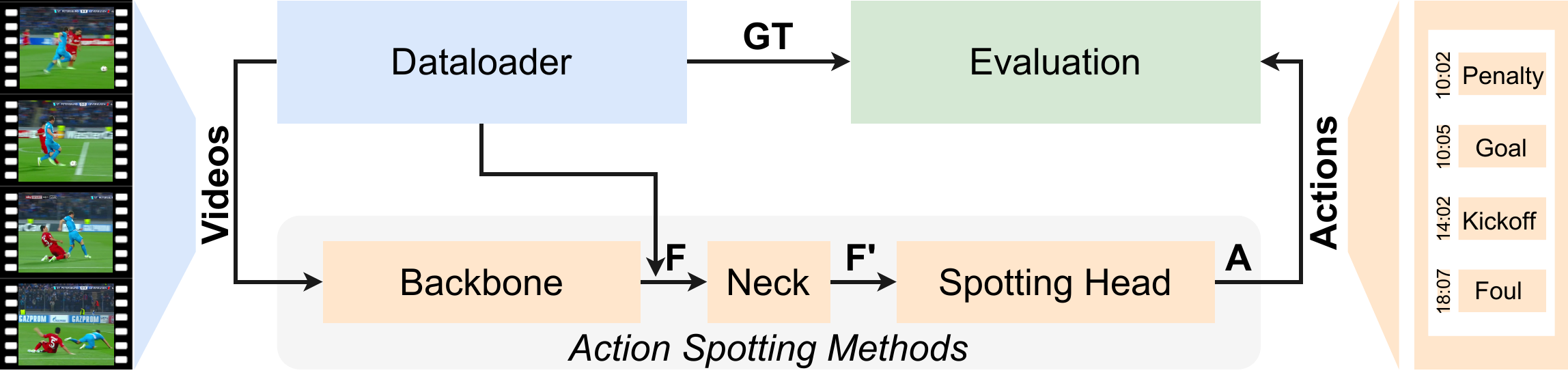}
    \caption{
    \textbf{\LIB} contains methods for efficient data loading, modularized action spotting, and comprehensive evaluation. 
    The dataloader provides videos $V$ or frame features $F$.
    The backbone converts the videos $V$ into frame features $F$. 
    The neck lifts the frame features $F$ into neck features $F'$, further processed by the spotting head that predicts the actions $A$.
    The evaluation compares the action predictions $A$ with the ground truth $GT$.
    }
    \label{fig:Pipeline}
\end{figure*}

\subsection{Definitions}

Action spotting in sports videos is the task of identifying and anchoring temporally specific actions semantic within a video.
Let \( V \) be a video with frames \( i_1, i_2, ..., i_T \), where \( T \) is the total number of frames. 
The objective is to detect the set of \( n \) actions \( A = \{a_1, a_2, ..., a_n\} \) in \( V \), each associated with a timestamp \( t_i \).
%
% \mysection{Action Spotting algorithm}
Formally, an action spotting algorithm is defined as a function \( \mathcal{F}: V \rightarrow A \times T \), mapping the video \( V \) to a set of action-timestamp pairs \( (a_i, t_i) \). 
This function aims to accurately align the predicted timestamps with the actual occurrence times of actions in the video.
The algorithmic structure can be segmented into three primary components: the backbone, the neck, and the head. 

\mysection{Backbone.}
The backbone is the fundamental part of the action spotting algorithm, primarily responsible for feature extraction. 
Given a video \( V \) with frames \( i_1, i_2, ..., i_T \), the backbone processes these frames to extract a set of features \( F = \{f_1, f_2, ..., f_T\} \). 
Formally, this can be represented as:
\[ F = Backbone(i_1, i_2, ..., i_T)\comma  \]
where \( F \) is the feature set corresponding to the frames in \( V \).

\mysection{Neck.}
The neck serves as an intermediate processing layer that refines and transforms the features \( F \) extracted by the backbone. 
Its purpose is to enhance the feature representation for more effective action spotting. The processed features \( F' \) can be expressed as:
\[ F' = Neck(F) \comma  \]
where \( F' \) are the neck features input to the spotting head.

\mysection{Head.}
The head is the final component of the algorithm, responsible for action identification and timestamp assignment. 
It utilizes the refined features \( F' \) to identify actions \( A \) and their corresponding timestamps \( T \). This can be formulated as:
\[ (A, T) = Head(F') \comma  \]
where \( A \) is the set of detected actions and \( T \) their timestamps.

Together, these components form the complete action spotting algorithm, where the backbone extracts raw features from video frames, the neck refines these features, and the head utilizes the refined features to detect and timestamp actions.
The pipeline is illustrated in Figure~\ref{fig:Pipeline}.
%This structured approach facilitates the effective and precise localization of actions within sports videos.

\subsection{Action Spotting Algorithms}

The \LIB library incorporates a variety of modular action spotting algorithms, each consisting of a backbone, neck, and head, to address different aspects of sports video analysis. The unified framework of \LIB is designed to operate with these algorithms,
% These algorithms are designed to operate within the unified framework of \LIB, ensuring compatibility and efficiency. 
Here, we detail some of the key action spotting algorithms already integrated into \LIB.

\mysection{Temporally-Aware Learnable Pooling~\cite{Giancola2021Temporally}.}
\textit{\underline{Backbone}}: The features are pre-extracted offline using a CNN (\eg ResNet), focusing on detailed spatial analysis of video frames.
\textit{\underline{Neck}}: It applies learnable and temporal pooling to these features, emphasizing temporal dynamics and relevance to action spotting. The learnable aspect is based on feature clustering (\eg NetVLAD), while the temporal aspect learns specific pooling before and after the action of interest.
\textit{\underline{Head}}: It processes further the pooled features with linear layers to predict the actions and their timestamps as a clip classification task.

\mysection{Context-Aware Loss Function (CALF)~\cite{Cioppa2020AContextaware}.}
\textit{\underline{Backbone}}: CALF utilizes the same pre-extracted features based on deep CNNs to learn spatial frame features.
\textit{\underline{Neck}}: The segmentation module learns the context of the action to spot, enhancing the features with semantic information around the actions.
\textit{\underline{Head}}: CALF employs the context-enhanced features from the neck in its spotting module for precise action localization regression for action candidates within the video clips.

\mysection{Precise Temporal Spotting (PTS)~\cite{Hong2022Spotting}.}
\textit{\underline{Backbone}}: PTS learns spatio-temporal features end-to-end with the remaining of the architecture. It typically relies on lighter spatial CNN such as RegNetY with intermediate temporal aggregation with GSM or TSM. 
\textit{\underline{Head}}: PTS does not contain any neck, it directly processes the backbone features in its recurrent spotting head (\eg GRU), which achieves a precise localization of the actions through a frame classification task.

These algorithms have been incorporated in \LIB, demonstrating the library's capability to cater to diverse needs in sports video analysis.
Through the systematic integration of these diverse algorithms, \LIB offers a robust and flexible solution for action spotting, adaptable to the specific requirements of different sports and analytical objectives.

% The action spotting algorithms are split between backbone, neck and head

% \mysection{Backbones}
% ResNet - RegNet

% % # From torchvision
% % 'rn18',
% % 'rn18_tsm',
% % 'rn18_gsm',
% % 'rn50',
% % 'rn50_tsm',
% % 'rn50_gsm',

% % # From timm (following its naming conventions)
% % 'rny002',
% % 'rny002_tsm',
% % 'rny002_gsm',
% % 'rny008',
% % 'rny008_tsm',
% % 'rny008_gsm',

% % # From timm
% % 'convnextt',
% % 'convnextt_tsm',
% % 'convnextt_gsm'

% \mysection{Neck}

% Pooling layers (Max - Avg - NetVLAD - NetRVLAD)

% Temporally-aware versions 

% \mysection{Head}

% Projection from 

% Linear layers

% Modules are handled with pytorch lightning.

% \mysection{Action Spotting Models}

% Feature based:
%  - Learnable pooling
%  - Temporally Aware Learnable pooling
%  - CALF

% End-to-End:
%  - PTS (No Neck)

\subsection{Data Handling and Processing}

In \LIB, the data handling and processing are paramount to ensure reproducible performance of action spotting algorithms.
The library accommodates two primary types of data loaders: feature-based and video-based, gathered into a unified action spotting dataset format.

\mysection{Feature-Based Data Loader.}
This loader is designed for scenarios where features are pre-extracted and stored separately from the raw video data. 
It is optimized for quick access and loading of these features, facilitating efficient processing in use cases where the action spotting algorithms rely on pre-computed feature sets. 
The feature-based data loader streamlines the workflow by directly ingesting these features into the action spotting models, bypassing the need for a backbone and thus speeding up the analysis process.

\mysection{Video-Based Data Loader.}
The video-based data loader, on the other hand, is tailored for processing raw video files directly, extracting features online as part of the action spotting pipeline. 
\LIB leverages both the OpenCV video reader and the NVIDIA DALI (Data Loading Library) for this purpose, an efficient video data loader that significantly accelerates the preprocessing and feature extraction stages. 
DALI optimizes the data pipeline, handling tasks like decoding, cropping, and resizing the video directly on the GPU, which minimizes the latency and computational overhead associated with these operations. 
The DALI-integrated video-based data loader in \LIB streamlines the training process by efficiently handling large-scale video datasets, ensuring fast and scalable video processing.
% suitable for high-throughput demands.
% This approach is particularly beneficial for faster video dataloading, hence algorithm training, directly from videos.
% The video-based data loader with DALI integration stands out for its ability to manage high-throughput video data, maintaining a balance between performance and computational efficiency. 
% By utilizing DALI, \LIB ensures that video processing is both fast and scalable, capable of supporting the intensive demands of training on large scale video daset.

% In summary, \LIB's data handling and processing capabilities, through its feature-based and video-based data loaders, provide a flexible and efficient framework for action spotting. 
% Whether leveraging pre-extracted features for speed or processing raw video content for flexibility, \LIB ensures that the data is prepared and optimized for the most efficient training of the action spotting algorithms.

\mysection{Unified action spotting dataset format.}
This new dataset format is a core aspect of \LIB, and is a novelty for the field of action spotting.
It is designed to harmonize the input data structure across various action spotting datasets, and encapsulates the essential elements of action spotting.
Inspired by the COCO format for image understanding, it is a single JSON file that includes the path of all videos or features, along with their annotations and associated metadata.
We envision this format to facilitate the integration of new datasets in \LIB, enabling easier training, fine-tuning, testing, and deployment on new domains.
% be widely used in action spotting, 
% such as temporal markers, action labels, and associated metadata—into a cohesive structure. 
% It enables the feature-based data loader to efficiently access pre-extracted features, while the video-based data loader can process raw videos to extract these features dynamically. 
% This unified format not only simplifies the integration of diverse data sources but also streamlines the workflow, allowing for seamless transition between feature extraction and action spotting phases.

In summary, \LIB's data handling, with its feature-based and video-based data loaders, as well as its unified action spotting dataset JSON format, offers a streamlined and adaptable framework for efficient action spotting training on novel domain, optimizing data preparation whether using pre-extracted features or raw video content.

% The Data Loading module in \LIB serves as the foundation for processing sports videos, handling the ingestion and initial preparation of video content for action spotting. This module is designed to efficiently manage the input of various video formats, extracting frames and normalizing them to a consistent size and format suitable for analysis. It ensures that the video data fed into the action spotting models is of high quality and uniformity, optimizing the performance of subsequent processing stages. By automating these preliminary steps, \LIB streamlines the workflow, allowing researchers to focus on the analytical aspects of action spotting without concerns about data compatibility or preprocessing issues.

% Numpy for pre-extracted features

% OpenCV for videos

% DALI for videos

\subsection{Evaluation metrics}

In \LIB, the evaluation framework for action spotting algorithms focuses on measuring their effectiveness and accuracy using two main metrics: loose and tight average mean Average Precision (mAP).

\mysection{Loose Average mAP:} This metric assesses the algorithm's performance with a lenient approach, allowing predictions to be considered correct if they fall within a broad temporal window around the actual event, ranging from $5$ to $60$ seconds. It gauges the algorithm's general ability to detect actions without requiring pinpoint temporal accuracy.

\mysection{Tight Average mAP:} Contrasting with the loose approach, the tight average mAP enforces a stringent temporal matching criterion, where predictions must closely align with the ground truth action timestamps, within a $1$ to $5$ second tolerance. This metric evaluates the algorithm's precision in accurately localizing actions in time.

These two mAP metrics provide a comprehensive view of an action spotting algorithm's performance in \LIB, balancing between general detection capabilities and precise temporal localization. Through this evaluation approach, \LIB ensures a robust assessment of action spotting algorithms, facilitating the refinement and enhancement of their performance in sports video analysis.

% The Evaluation module in \LIB is crucial for validating the effectiveness and accuracy of the action spotting algorithms. It employs a comprehensive set of metrics and testing methodologies to assess the performance of the models under various conditions and datasets. This module not only quantifies the success of action spotting in terms of precision, recall, and computational efficiency but also provides insights into the models' behavior, identifying strengths and areas for improvement. Through rigorous evaluation, \LIB ensures that the action spotting algorithms meet the high standards required for reliable sports video analysis, facilitating continuous refinement and optimization of the models.

% Average-mAP (tight + loose)

% Data Input: Describe the types of data \LIB can handle, including video formats, resolution, and source variations.
% Preprocessing: Explain the preprocessing steps, such as frame extraction, normalization, and augmentation, to prepare data for action spotting.
% Post-processing: Detail any post-processing steps, like action classification aggregation, temporal localization refinement, or results visualization.

% \subsection{Performance Optimization}

% Faster than OpenCV?

% Efficiency Improvements: Outline the measures taken to optimize the library’s performance, including computational efficiency, memory management, and real-time processing capabilities.
% Scalability: Discuss how \LIB is designed to scale with increasing data volumes or computational demands.

\subsection{User-friendly Tools and Module}

\LIB utilizes the PyTorch Lightning framework to offer streamlined tools for training, inference, and evaluation, enhancing the ease of use and efficiency in training action spotting algorithms.
The \textbf{training tools} facilitate model training with automated batch processing and checkpointing, allowing for flexible and efficient model development.
The \textbf{inference tools} enable quick and accurate action detection in videos, optimized for both real-time and batch-processing environments.
The \textbf{evaluation tools} provide detailed performance metrics to assess and compare the effectiveness of action spotting models within \LIB.
These tools, integrated in \LIB with PyTorch Lightning, simplify the deep learning workflow, making advanced action spotting accessible and manageable for users across the sports analytics field.

% Pytorch Lightning

% We also Tools for train/infer/evaluate

% `train()`

% `infer()`

% `evaluate()`

% Interface Design: Describe the user interface of \LIB, focusing on how it facilitates easy access to its functionalities, including algorithm selection, parameter tuning, and results visualization.
% Documentation and Support: Highlight the availability of comprehensive documentation, tutorials, and user support to aid in the effective use of \LIB.

% \subsection{Validation and Testing}

% Tested on SoccerNet-v2, got same results.

% Testing Framework: Explain the approaches used for testing and validating the library, including unit tests, integration tests, and user acceptance testing.
% Dataset Utilization: Describe the datasets used for testing and validating the library's action spotting capabilities, emphasizing the variety and scope of the data.
\section{Experiments / Benchmark}

% \subsection{Experimental Setup}
We validate the reproducibility of our \LIB implementation with respect to established methods.
% We assess the performance of our \LIB implementations, matching performance against established %methods.
% We evaluate our \LIB implementations on the widely-used SoccerNet-v2 action spotting dataset, %matching performance against baseline methods.
% We test the methods developed in \LIB on the popular action spotting dataset SoccerNet-v2, to validate the performance of our implementation with respect to the original implementations. 

\mysection{Dataset.} We focus our experiments on the SoccerNet-v2 action spotting dataset~\cite{Deliege2021SoccerNetv2}, as it is the most widely used dataset in the literature. 
Yet, more datasets can be tested with our library.
SoccerNet-v2 provides videos for $500$ football games from the big five European leagues, fully annotated with more than $110{,}458$ temporally-anchored actions, from $17$ common action classes. 
We refer to the SoccerNet-v2 paper~\cite{Deliege2021SoccerNetv2} for a more detailed description of the dataset.

\mysection{Methods.} 
For the feature-based methods, we focus our experiments on the pre-extracted ResNet-152 features reduced with a PCA at a dimension of 512, provided with the SoccerNet-v2 dataset~\cite{Deliege2021SoccerNetv2}. 
For the \underline{\textit{Temporally-Aware Learnable Pooling}} methods~\cite{Giancola2021Temporally}, we test the non-parametric pooling necks (MaxPooling, AvgPool), the learnable pooling necks (NetVLAD++, NetRVLAD++) and the temporally-aware pooling necks (Maxpool++, Avgpool++, NetVLAD++, NetRVLAD++). For the cluster-based approaches, we set the number of clusters to $64$. The head is a single linear layer that projects the neck features to a dimension equal to the class number, followed by a softmax activation for the logits. We keep the training parameters similar to the original implementation.
For the \underline{\textit{Context-Aware Loss Function (CALF)}} method~\cite{Cioppa2020AContextaware}, we define the segmentation module as the neck, and the spotting head as the head. We keep the training parameters and the temporal parameters similar to the original implementation.
For the end-to-end approach \underline{\textit{Precise Temporal Spotting (PTS)}}~\cite{Hong2022Spotting}, we use the 224p videos at 25fps as input. The backbone processes the frames at 2fps with RegNetY (RNY) and a GSM temporal shift. PTS does not have any neck, and we select the GRU head to accumulate the temporal information and predict the class activations. We keep the training parameters similar to the original implementation meant for the SoccerNet-v2 dataset.

% \mysection{Hardware and Software Configuration:} Describe the computing environment, including hardware specifications and software versions, to ensure reproducibility.

% \mysection{Library Configuration:} Detail the specific setup of \LIB used during the experiments, including any relevant parameters and settings.

\mysection{Results.}
We present in Table~\ref{tab:main_results} the results of our experiments.
% We can see that ...
We reproduce the same performances as the original implementation in terms of loose and tight Average-mAP.
We report the training time (per epoch between parenthesis), as well as the time for inference of the complete testing set of SoccerNet-v2 composed of $100$ videos of $45$min.
The timing for all methods was estimated on 1 GPU RTX1080, 2 CPU cores and 32GB RAM, except for the PTS that required 4 GPU RTX A500 and 80GB of RAM. 

% Performance Analysis: Present the results of the experiments, including statistical analyses, graphs, and charts that illustrate the performance of \LIB across different datasets and in comparison to other methods.
% Discussion of Findings: Analyze the results, highlighting the strengths and limitations of \LIB as demonstrated by the experiments

\begin{table}[ht]
    \centering
    \caption{
    \textbf{\LIB: Reproduced Results.}
    }
\resizebox{\linewidth}{!}{%
    \begin{tabular}{l||c|c|c|c}
\toprule
\multirow{2}{*}{Method} &  \multicolumn{2}{c|}{Average-mAP} &  \multicolumn{2}{c}{Timing (sec)}  \\ \cmidrule{2-5}
           & loose       & tight     & Training  & Testing \\ \midrule
Maxpool~\cite{Giancola2021Temporally}    & 20.8        & ~2.1  &   1268 (7)    &  ~118 \\ % (18.6)
Avgpool~\cite{Giancola2021Temporally}    & 29.4        & ~2.3  &   ~886 (7)    &  ~118 \\ % (23.7)
NetVLAD~\cite{Giancola2021Temporally}    & 46.3        & ~4.4  &   1328 (7)    &  ~119 \\ % (31.4)
NetRVLAD~\cite{Giancola2021Temporally}   & 44.8        & ~4.8  &   1371 (7)    &  ~118 \\ \midrule
Maxpool++~\cite{Giancola2021Temporally}  & 32.2        & ~5.3  &   1172 (7)    &  ~118 \\ 
Avgpool++~\cite{Giancola2021Temporally}  & 40.8        & ~6.9  &   ~974 (7)    &  ~118 \\ 
NetVLAD++~\cite{Giancola2021Temporally}  & 51.5        & ~8.1  &   1185 (8)    &  ~120 \\ % (50.7)
NetRVLAD++~\cite{Giancola2021Temporally} & 49.8        & ~8.3  &   1147 (8)    &  ~118 \\ \midrule
CALF~\cite{Cioppa2020AContextaware}      & 39.5        & 12.4  &   1639 (4)    &  ~~26 \\ \midrule % 39.45 (40.7)
PTS (RNY002)~\cite{Hong2022Spotting}              & 70.4        &    63.9  & 140700 (1200) &  1500 \\ 
PTS (RNY008)~\cite{Hong2022Spotting}              & 72.2        &    66.2  & 249780 (1500) &  1800 \\ 
\bottomrule
    \end{tabular}
    }
    \label{tab:main_results}
\end{table}

\section{Discussions}

\mysection{Performances reproduction.}
\LIB successfully meets its primary goal of reproducing the performance metrics documented in existing literature, demonstrating its capability and reliability in the domain of sports video analysis. 

\mysection{DALI: accelerated video loading.}
Integrating DALI for video loading in \LIB improved the processing of the video data, overcoming the inefficiencies of traditional CPU-based loading methods like OpenCV video and PyTorch image data loaders. 
This shift to GPU processing with DALI results in significant time improvement, streamlining \LIB’s data pipeline. 
Compared to OpenCV video data loaders, best cases result in around 33\% time improvement for the training steps and 50\% improvement for each epoch's validation steps.
Furthermore, compared to the PyTorch image data loader, we remove the need of pre-extracting frames from videos.
Pre-extracting frame typically requires a significant extra volume space to store this data, typically around 10 fold, and presents a reading bottleneck due to a large number of file access.
By adopting DALI, \LIB enhances pre-processing efficiency and sets a new standard for speed, storage, and performance in sports video analysis, paving the way for more sophisticated and expansive video pre-processing for action spotting research, including larger resolution and higher frame rate.

\mysection{JSON-based action spotting dataset format.}
The JSON-based action spotting dataset format in \LIB marks a significant advance in standardizing data handling for sports analytics, aligning with algorithm needs by organizing timestamps, action labels, and metadata cohesively. 
This format enhances interoperability and eases integration with analytics tools, streamlining dataset preparation and parsing for efficient model training and evaluation. 
Standardization with \LIB's JSON format boosts consistency in video analysis, supporting more scalable and collaborative research efforts, and facilitating innovation in sports analytics.

\section{Conclusion}

\LIB marks a novel development in sports video analysis by introducing the first Python library to consolidate major action spotting algorithms into a modular framework, fostering future algorithmic development. 
Its efficient video data loader, powered by DALI, streamlines video pre-processing, facilitating rapid training of end-to-end algorithms. 
Furthermore, the introduction of a novel JSON-based action spotting dataset format with \LIB enhances the adaptability and application of its algorithms across various datasets and video analyses. 
We envision \LIB as the one-stop Python library for action spotting algorithms, that will centralize the efforts of the sports analytics community into an easy-to-use modular library, accelerating the development and deployment of these tools.

% We believe \LIB lay a solid foundation for ation spotting algorithmic development, setting a clear trajectory for its evolution as a central tool in sports analytics, with a vision to expand its capabilities and influence in transforming the landscape of sports video analysis

% These contributions not only underscore \LIB’s significance in advancing sports analytics but also pave the way for its broader adoption and impact in the field.
\section*{Acknowledgment}
A. Cioppa is funded by the F.R.S.-FNRS. This work was partly supported by the King Abdullah University of Science and Technology (KAUST) Office of Sponsored Research through the Visual Computing Center (VCC) funding and the SDAIA-KAUST Center of Excellence in Data Science and Artificial Intelligence (SDAIA-KAUST AI).
Bruno Cabado wish to thanks the Axencia Galega de Innovación the grant received through its Industrial Doctorate program (23/IN606D/2021/2612054). CITIC is funded by Xunta de Galicia (ED431G 2019/01) and ERDF funds.

\bibliographystyle{ieeetr}
\bibliography{all-arxiv}
%bib/transferanalysis}

% that's all folks
\end{document}